\documentclass[sn-mathphys-num]{sn-jnl}

\usepackage{graphicx}%
\usepackage{multirow}%
\usepackage{amsmath,amssymb,amsfonts}%
\usepackage{amsthm}%
\usepackage{mathrsfs}%
\usepackage[title]{appendix}%
\usepackage{xcolor}%
\usepackage{textcomp}%
\usepackage{manyfoot}%
\usepackage{booktabs}%
\usepackage{algorithm}%
\usepackage{algorithmicx}%
\usepackage{algpseudocode}%
\usepackage{listings}%

\usepackage{hyperref}

\usepackage{pifont}
\newcommand{\cmark}{\ding{51}}
\newcommand{\xmark}{\ding{55}}
\usepackage{multirow}

\usepackage{color, colortbl}
\definecolor{Gray}{gray}{0.9}

\usepackage{comment}

\theoremstyle{thmstyleone}%
\theoremstyle{thmstyletwo}%

\theoremstyle{thmstylethree}%

\begin{document}

\title[Article Title]{Swapped Logit Distillation via Bi-level Teacher Alignment}

\author[1]{\fnm{Stephen Ekaputra} \sur{Limantoro}}\email{stephen.ee08@nycu.edu.tw}

\author[1]{\fnm{Jhe-Hao} \sur{Lin}}\email{jimmylin0979.11@nycu.edu.tw}

\author[2]{\fnm{Chih-Yu} \sur{Wang}}\email{eric\_wang@cyberlink.com}

\author[2]{\fnm{Yi-Lung} \sur{Tsai}}\email{bruceyl\_tsai@cyberlink.com}

\author[1]{\fnm{Hong-Han} \sur{Shuai}}\email{hhshuai@nycu.edu.tw}

\author[1]{\fnm{Ching-Chun} \sur{Huang}}\email{chingchun@nycu.edu.tw}

\author[3]{\fnm{Wen-Huang} \sur{Cheng}}\email{wenhuang@csie.ntu.edu.tw}

\affil*[1]{\orgdiv{Department of Electrical Engineering and Computer Science}, \orgname{National Yang Ming Chiao Tung University}, \orgaddress{\city{Hsinchu} \postcode{300093}, \country{Taiwan}}}

\affil[2]{\orgname{Cyberlink}, \orgaddress{\city{New Taipei City} \postcode{231}, \country{Taiwan}}}

\affil[3]{\orgdiv{Department of Computer Science and Information Engineering}, \orgname{National Taiwan University}, \orgaddress{\city{Taipei} \postcode{10617}, \country{Taiwan}}}


\abstract{Knowledge distillation (KD) compresses the network capacity by transferring knowledge from a large (teacher) network to a smaller one (student). It has been mainstream that the teacher directly transfers knowledge to the student with its original distribution, which can possibly lead to incorrect predictions. In this article, we propose a logit-based distillation via swapped logit processing, namely Swapped Logit Distillation (SLD). SLD is proposed under two assumptions: (1) the wrong prediction occurs when the prediction label confidence is not the maximum; (2) the “natural” limit of probability remains uncertain as the best value addition to the target cannot be determined. To address these issues, we propose a swapped logit processing scheme. Through this approach, we find that the swap method can be effectively extended to teacher and student outputs, transforming into two teachers. We further introduce loss scheduling to boost the performance of two teachers' alignment. Extensive experiments on image classification tasks demonstrate that SLD consistently performs best among previous state-of-the-art methods. Codes are available at \href{https://github.com/StephenEkaputra/Swapped-Logit-Distillation}{GitHub}.

}


\maketitle

\section{Introduction}\label{sec1}

\begin{figure}
\centering
\includegraphics[width=0.8\columnwidth]{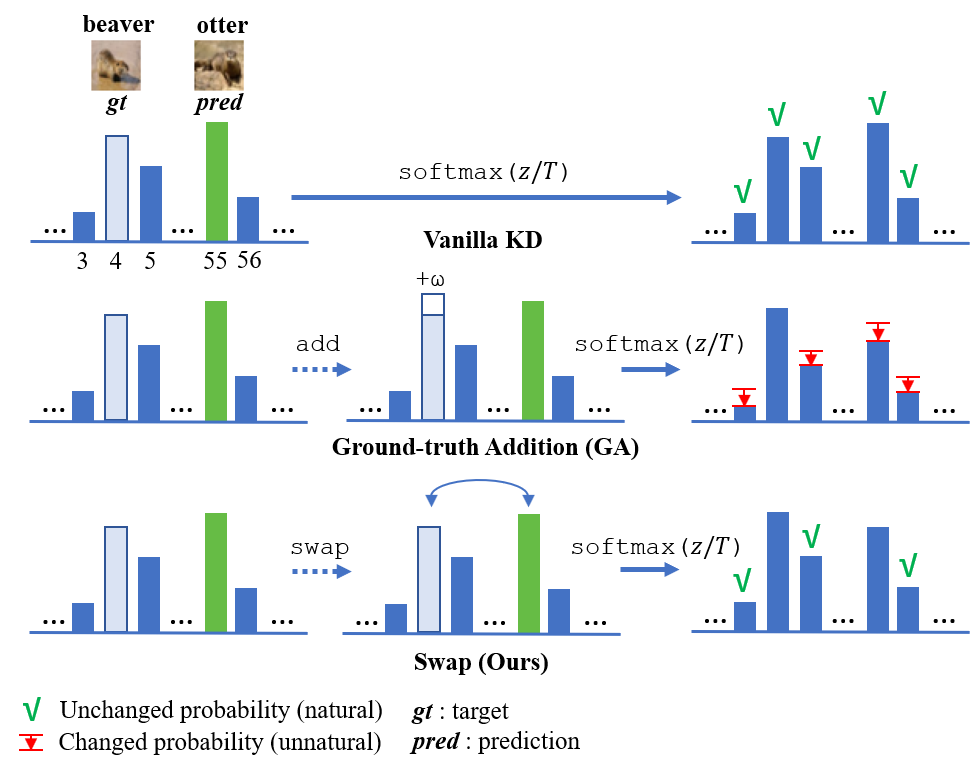}
\caption{Illustration of intuitive logit processing schemes with a false prediction case.}
\label{fig:logit_processing}
\end{figure}

In the last few decades, computer vision with Deep Neural Networks has made significant advancements in various applications, such as image classification \cite{resnet,vit}, object detection \cite{fasterrcnn, dq_detr}, visual tracking \cite{s2siamfc}, fashion in vision \cite{fashion_survey, fashion_on, fashion_recommendation}, hand gesture recognition \cite{lared, hagrid}, facial age estimation \cite{facial_estimation}, micro-expression recognition \cite{micro_expression}, and scene generation \cite{llm_scene}. However, robust deep learning models generally align with high computational and storage costs, which are unreliable in practical scenarios like mobile device applications. To reduce the model capacity, three common approaches can be employed to develop a lightweight model \cite{lightweight_survey}, \textit{i.e.,} neural network architecture design, compression method, and hardware acceleration. Among those, knowledge distillation (KD) \cite{kd} is an effective compression technique where a teacher model (large) transfers the knowledge to a student model (small) via reducing the KL divergence from the logits. This technique can improve the student model's performance without any extra cost.

KD typically falls into two main categories, \textit{e.g.}, logit \cite{kd,dkd} and feature distillation methods \cite{fitnets,ofd,reviewkd}. These categories represent two aspects of knowledge transfer from a teacher to a student model. The logit distillation transmits soft target information from teacher to student on the logit level. On the other hand, feature distillation aims to transfer feature representation from the intermediate layer. Compared with the former logit distillation \cite{kd}, the performance of feature distillation is generally superior on various tasks \cite{ofd,crd,reviewkd}. However, the feature distillation method has two main drawbacks, such as 1) extra training time for distilling deep features and 2) unavailability of core architecture and intermediate layers in certain deep learning applications \cite{mlkd}. In the later research, numerous researchers \cite{dkd,mlkd,normkd} revisited the logit-based distillation method in which modified distillation loss is introduced to help the student model effectively leverage the knowledge from the teacher model, yielding distillation results that are comparable to or even better than those achieved through feature-based methods.

\begin{figure}[ht]
\includegraphics[scale=0.45]{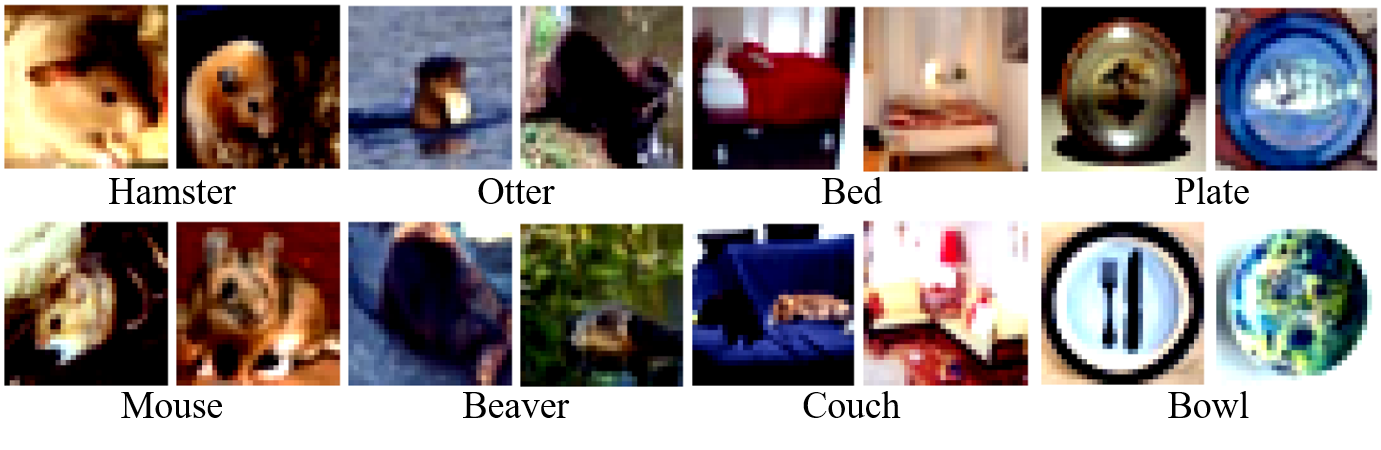}
\centering
\caption{Various samples (animals and objects) that KD and MLKD misclassify are correctly classified by SLD. The top and bottom images display high feature similarities.}
\label{fig:visual}
\end{figure}

Most existing logit-based KD approaches still assume that the student should learn from the teacher directly, neglecting the possibility of transferring misclassified logit information. This may occur when the target is not the highest one. An intuitive reason is that the target has similar features to the non-target, as indicated by high confidence. As shown in Fig. \ref{fig:logit_processing}, class 4 (beaver) and class 55 (otter) belong to the same superclass “aquatic mammals” that share similar features, such as color, texture, and shape, which may mislead the learning of the student model. In fact, the correct prediction is class 4 (beaver). We can also see that after preprocessing with the softmax function, the vanilla KD keeps the distribution natural, but the prediction is false. On the other hand, we introduce Ground-truth addition (GA) by adding the target class with a fixed small value percentage producing a correct prediction, but it changes the distribution arbitrarily, which may lead to information loss. In contrast, our approach corrects the prediction and keeps the distribution “natural” simultaneously. We demonstrate our results on various samples in Fig. \ref{fig:visual}.

To this end, we propose the Swapped Logit Distillation (SLD) reforming the logit distillation via the swap logit mechanism. Concretely, the target is swapped with the non-target with the highest probability. As a result, the target will be increased, and the non-target with maximum confidence will be reduced in a failed prediction case, effectively killing two birds with one stone. Through this approach, the student model absorbs the “dark knowledge” with the true target in a more natural way. We then hypothesize if the swap processing effectively works with teacher logit, it can also work with student logit since the distribution is from the same source. Henceforth, the swap method is applied to both teacher and student logits. However, to prevent conflicts in teachers' alignment, we introduce loss scheduling, where the pseudo-teacher is incorporated after the developed student mimics the original teacher. This will advance the use of the pseudo-teacher. We subsequently adopt multiple temperature scaling from MLKD \cite{mlkd} alongside our swap method to extract richer information across diverse scales. As a result, SLD aligning all elements with merely logit information demonstrates its effectiveness over state-of-the-art methods.

The main contributions are summarized as follows:
\begin{itemize}
\item We discover and discuss the shortcomings of mainstream logit-based KD methods and the effects of various logit processing in terms of naturality. This leads to the transfer of imprecise information and information loss from the teacher to the student models. 
\item We propose a simple and effective method, \textit{i.e.}, SLD. The misclassified prediction is swapped with the highest confidence to correct the prediction and keep the distribution “natural”. The swap method can be effectively applied to teacher and student outputs, allowing the student to learn from the teacher and the pseudo-teacher. To avoid the learning conflict of pseudo-teacher with the teacher, we introduce loss scheduling and show its benefits on teacher integration.
\item Experimental results manifest that SLD works well when combined with other distillation methods, including both feature and logit-based distillation, demonstrating the generalizability of the student model.
\end{itemize}

\section{Related work}
\subsection{Logit-based Distillation}
The earliest knowledge distillation (KD) was introduced by Hinton et al. \cite{kd} as a logit distillation method. KD aims to improve the performance of the student network by transferring soft labels from the teacher network. Previous works of logit distillation introduced a mutual-learning manner \cite{dml} and teacher assistant module \cite{takd}. DKD \cite{dkd} decouples the classical KD loss into the target and non-target parts and calculates the losses separately. In the later work, MLKD \cite{mlkd} proposes prediction augmentation with strong augmentation and multi-level alignment to reduce the divergence at the instance, class, and batch levels with multiple different losses. Notably, our approach is more efficient and simple than MLKD since we focus on instance-level relations, disregarding the computational complexity of the Gram matrix and potential issues associated with balancing multiple distinct losses. Besides, the existing logit-based methods focus on the native logit information, which may contain errors in transferring knowledge. 

\subsection{Feature-based Distillation}
The feature distillation technique is introduced to enhance the effectiveness of knowledge distillation. This approach involves the distillation of intermediate features along with the logit outputs. Several approaches \cite{ofd,kdab,fitnets} strive to reduce the divergence between the features of teacher and student models, compelling the student model to replicate the teacher model's features. Other methods \cite{rkd,crd,skd} transfer the correlation input of the teacher to the student. Most of the feature distillation methods achieve better performance due to their rich knowledge absorption from the teacher model. In contrast to the feature-based approach, this paper aims to merely use logit information and improve the performance of previous state-of-the-art works.

\section{Methodology}
First, we dive into the background of original knowledge distillation. We reconstruct logit distillation with a simple logit processing scheme to enhance the logit quality naturally. Then, a brief discussion is presented to demonstrate the effects of various logit processing schemes.

\begin{figure*}[ht]
\centering
\includegraphics[scale=0.50]{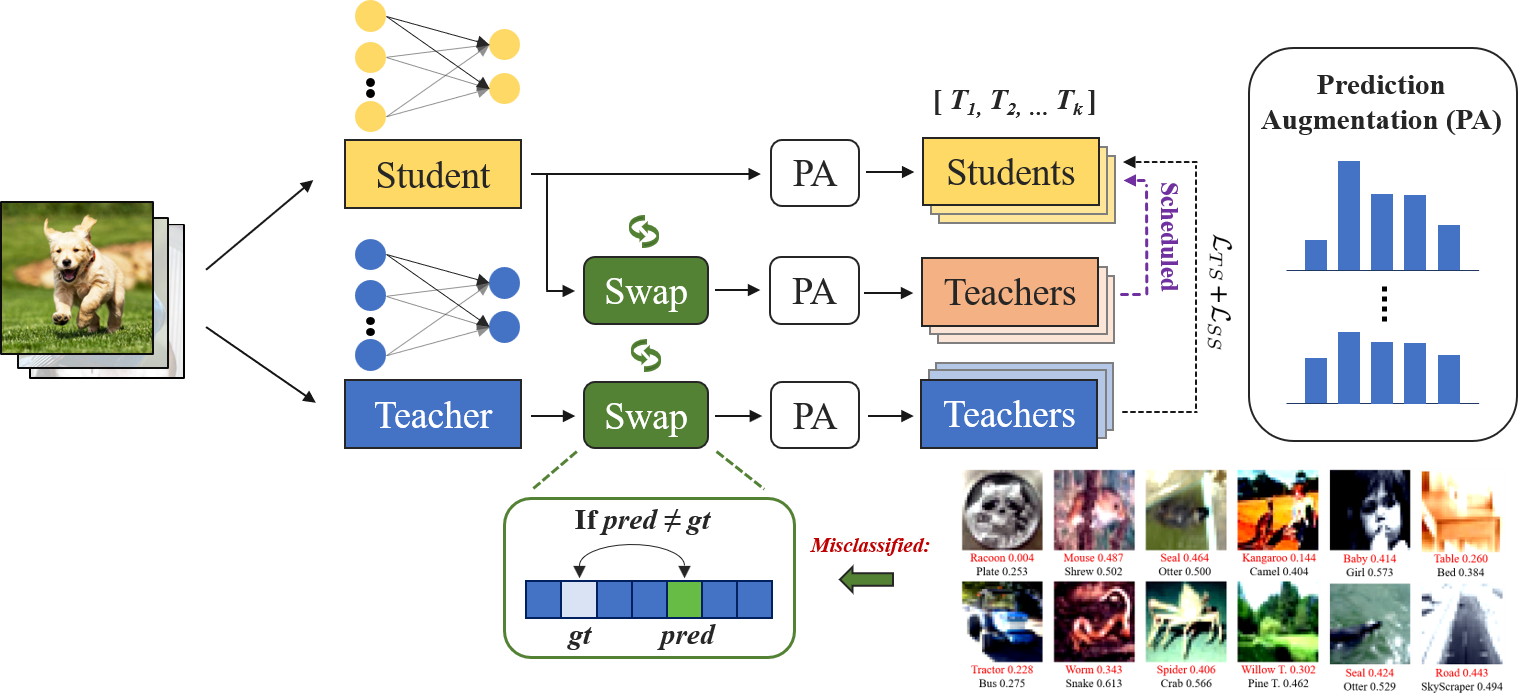}
\caption{Overview of SLD framework. After obtaining the logit outputs, swapping methods are applied to teacher logits, producing main teacher and pseudo-teacher logits. Loss scheduling is proposed to avoid teacher conflicts when both teachers are integrated. Prediction augmentation is used to generate outputs at various softness levels. Real misclassified samples of the teacher's prediction illustrate how a model prediction can mislead the student model's learning since the target and the prediction share similarities in color, shape, and textures (where the red text is the target, and the black one is the false prediction).}
\label{fig:framework}
\end{figure*}

\subsection{Background}
\noindent\textbf{Knowledge Distillation.}
According to the classical KD \cite{kd}, the softmax function plays an important role in converting the logit into a probability over classes, indicating how likely each class is for a given input ${z\in \mathbb{R}^{C}}$. The softmax function can be calculated by:

\begin{equation}
\begin{aligned}
\label{softmax}
p_{j} = \frac{exp{(z_{j}/T)}}{\sum^{C}_{c=1} exp{(z_{c}/T)}},
\end{aligned}
\end{equation}
where $z_{j}$ and $p_{j}$ represents the logit and probability output on the $j$-th class, respectively. $T$ is the temperature scaling to scale the smoothness of the distribution. To measure the match of the teacher and student's output, the Kullback Leibler (KL) divergence loss is adopted as follows:  

\begin{equation}
\begin{aligned}
\label{kdloss}
\mathcal{L}_{KD} = KL(p^{tea} || p^{stu}) = \sum^{C}_{j=1} p_{j}^{tea} log \left( \frac{p_{j}^{tea}} {p_{j}^{stu}} \right), 
\end{aligned}
\end{equation}
where $\mathcal{L}_{KD}$ is the KL loss, $p_{j}^{tea}$ and $p_{j}^{stu}$ are the teacher's and student's probability output on the $j$-th class, respectively.

\subsection{Swapped Logit Distillation}
In this section, we reformulate KD to SLD by coupling prediction augmentation and swapping mechanisms. SLD has three new main components compared with the previous works: teacher swap loss, student swap loss, and loss scheduling for teacher alignment. The overview of the overall framework is depicted in Fig. \ref{fig:framework}.

\noindent\textbf{Prediction Augmentation.}
To increase logit diversity, prediction augmentation \cite{mlkd} is adopted, where the prediction output is expanded into multiple ones with different temperature scales as it is proven to improve the generalization.

\begin{equation}
\begin{aligned}
\label{predictionaugmentation}
p_{i,j,k} = \frac{exp(z_{i,j}/T_{k})}{\sum^{C}_{c=1} exp(z_{i,c}/T_{k})},
\end{aligned}
\end{equation}
where $p_{i,j,k}$ is the probability output of $i$-th input on the $j$-th class. The temperature scaling $T_{k}$ is scaled into $[T_{1}, ..., T_{k}]$, where $k$ = 6. Note that the PA mechanism enables the swap mechanism to gain richer information on logit distillation at several softness levels. Compared with MLKD \cite{mlkd}, SLD merely focuses on the instance-level relations in knowledge transfer. For simplicity and compactness, the swap mechanism is proposed to replace the batch-level and class-level relations, and we demonstrate the effectiveness of SLD over the previous state-of-the-art methods in the experiment section.

\noindent\textbf{Teacher Swap.} 
It is common to process the logit directly with the softmax function to obtain the probability output. Meanwhile, the prediction of the teacher model is not always correct. When the teacher model's prediction is incorrect, the student model can be influenced by the teacher model's behavior with misleading information. The proposed method aims to correct the teacher model's prediction by swapping the ground truth and index with the highest confidence. As a result, the natural maximum limit of the teacher model's logit remains the same, and the prediction is correct simultaneously. We then minimize the KL divergence of the swapped teacher and student as follows:

\begin{equation}
\begin{aligned}
\label{teacherswap}
    p^{tea^{\prime}}_k= 
\begin{cases}
    swap(p^{tea}_k),& \text{if } \arg\max(p^{tea}_k)\neq t\\
    p^{tea}_k, & \text{otherwise}
\end{cases}
\end{aligned}
\end{equation}

\begin{equation}
\begin{aligned}
\label{teacherswaploss}
\mathcal{L}_{TS} = \sum^{K}_{k=1}KL(p^{tea^{\prime}}_k || p^{stu}_k)
\end{aligned}
\end{equation}
where $\mathcal{L}_{TS}$ is the teacher swap loss, $t$ is the target, $p^{tea^{\prime}}$ and $p^{stu}$ are the new probability output associated with swapped teacher and student, respectively, augmented through multiple temperatures $T_{k}$.

\noindent\textbf{Student Swap as Pseudo-teacher.} 
The next part of the proposed method is the student swap loss. The same swap mechanism is also applied to the student's logit, which becomes a pseudo-teacher. This is because the information output is different and independent from the teacher's. Our objective is to mitigate the divergence between the student and the pseudo-teacher. Therefore, the corresponding loss would be:

\begin{equation}
\begin{aligned}
\label{studentswap}
p^{stu^{\prime}}_k= 
\begin{cases}
swap(p^{stu}_k),& \text{if } \arg\max(p^{stu}_k)\neq t\\
p^{stu}_k,              & \text{otherwise}
\end{cases}
\end{aligned}
\end{equation}

\begin{equation}
\begin{aligned}
\label{studentswaploss}
\mathcal{L}_{SS} = \sum^{K}_{k=1}KL(p^{stu^{\prime}}_k || p^{stu}_k)
\end{aligned}
\end{equation}
where $\mathcal{L}_{SS}$ is the student swap loss and $p^{stu^{\prime}}$ is the new pseudo-teacher's probability output augmented by $T_{k}$.

\begin{figure*}[h]
\centering
\includegraphics[scale=0.4]{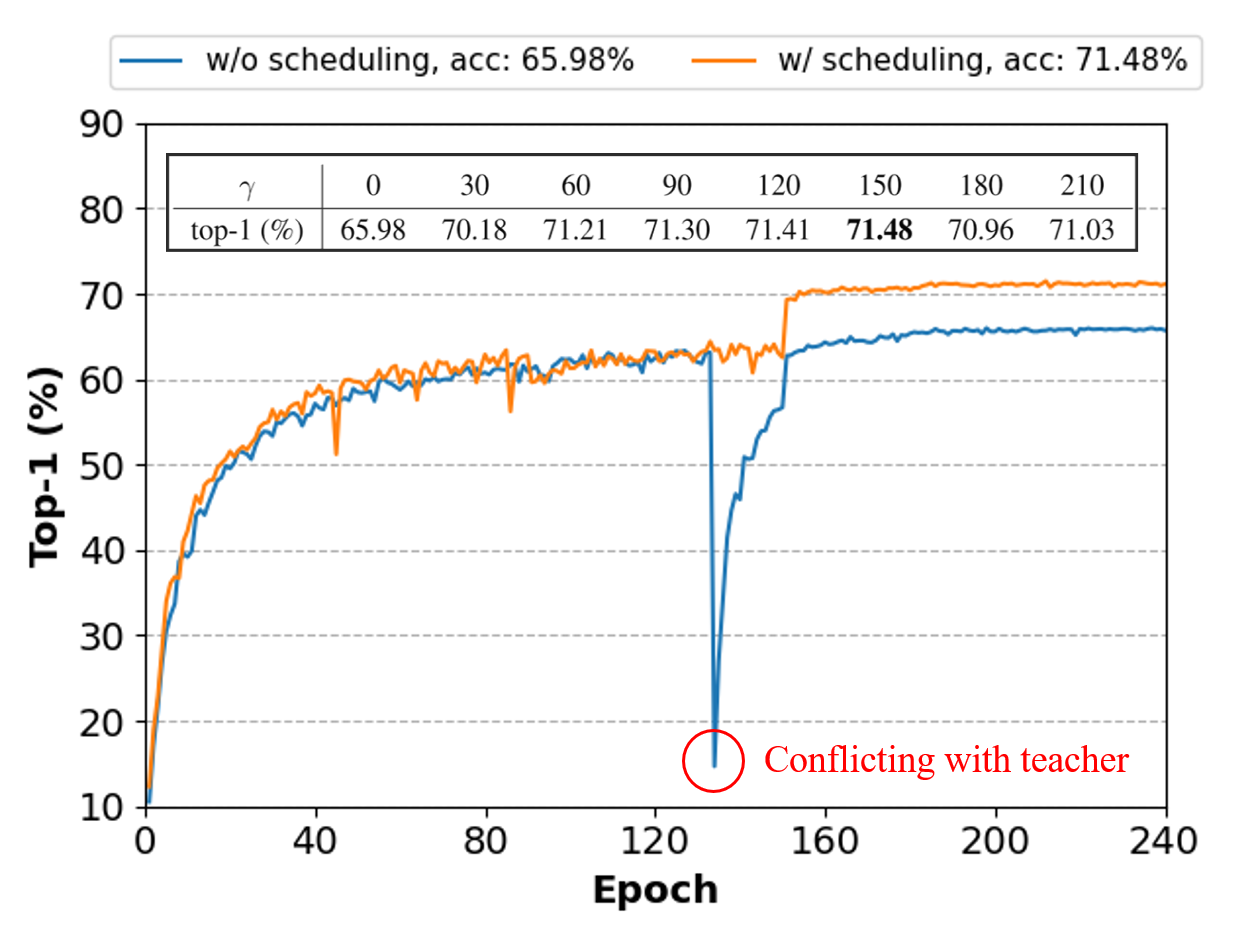}
\caption{Effects of scheduling on ResNet50 as a teacher and MobileNet-V2 as a student on CIFAR-100.}
\label{fig:scheduling}
\end{figure*}

\noindent\textbf{Loss Scheduling for Teacher Alignment.} 
Unlike the teacher model, which has been fully trained and produces a more stable distribution, the student model at the early epoch has not yet converged. This may cause conflict in soft-label learning between teacher output and pseudo-teacher output when they are integrated from the first training epoch. In Fig \ref{fig:scheduling}, we show an experiment in which bi-level teachers are integrated with an ablation study. To avoid alignment conflicts, we simply add the $\mathcal{L}_{SS}$ after a certain training epoch. It is written as: 

\begin{equation}
\begin{aligned}
\label{scheduleloss}
\mathcal{L}_{SS}= 
\begin{cases}
\mathcal{L}_{SS},& \text{if } epoch > \gamma\\
0,              & \text{otherwise}
\end{cases}
\end{aligned}
\end{equation}
where $\gamma$ denotes the scheduled epoch. Based on the ablation, the $\gamma$ is set after the first learning rate division to 30 and 150 on ImageNet and CIFAR-100 datasets, respectively.

\begin{algorithm}[t]
\caption{Pseudo code of SLD in a PyTorch-like style.}
\renewcommand{\arraystretch}{1.3}
\label{sld_algo}
\definecolor{codeblue}{rgb}{0.1,0.6,0.6}
\lstset{
backgroundcolor=\color{white},
basicstyle=\fontsize{7.2pt}{7.2pt}\ttfamily\selectfont,
columns=fullflexible,
breaklines=true,
captionpos=b,
commentstyle=\fontsize{7.2pt}{7.2pt}\color{codeblue},
keywordstyle=\fontsize{7.2pt}{7.2pt},
}

\begin{lstlisting}[language=Python,numbers=none]
# z_stu, z_tea: student, teacher logit outputs
# T = [T_1, T_2,...,T_k] temperature scaling
# z_stu_new, z_tea_new: new student, teacher logit outputs
# t: target, n: non-target with maximum confidence
# loss_ts, loss_ss: teacher swap, student swap loss
# gamma: scheduled epoch

z_tea_new = torch.clone(z_tea)
z_stu_new = torch.clone(z_stu)

# swap mechanism, Eqn.(4)&(6)
if z_tea[t] != max(z_tea): # teacher swap
    z_tea_new[t], z_tea_new[n] = z_tea[n], z_tea[t]
if z_stu[t] != max(z_stu): # student swap
    z_stu_new[t], z_stu_new[n] = z_stu[n], z_stu[t]

loss_sld = 0

# multi-alignment
for i in T:
    p_stu = F.softmax(z_stu / i)
    p_tea_1 = F.softmax(z_tea_new / i)
    p_tea_2 = F.softmax(z_stu_new / i) 

    loss_ts = F.kl_div(p_tea_1, p_stu)
    # loss scheduling
    if epoch > gamma:
        loss_ss = F.kl_div(p_tea_2, p_stu)
    else:
        loss_ss = 0
    loss_sld += (loss_ts + loss_ss)
    
\end{lstlisting}
\end{algorithm}

\noindent\textbf{Final Alignment.} 
In the final part, we integrate teacher and student swap loss to be $\mathcal{L}_{SLD}$ that can be rewritten as follows:

\begin{equation}
\begin{aligned}
\label{finalloss}
\mathcal{L}_{SLD} = \mathcal{L}_{TS} + \mathcal{L}_{SS}
\end{aligned}
\end{equation}

Two instance-level losses are fused together to assist the student model in mimicking bi-level teachers, including the teacher and the pseudo-teacher. By swapping the logit, the student model pays more attention to the target and less attention to the misclassified non-target. We provide the Pytorch-like style pseudo-code in Algorithm \ref{sld_algo}.

\begin{table*}[h]
\renewcommand{\arraystretch}{1.1}
\caption{Comparison of logit processing conducted on CIFAR-100 with ResNet32$\times$4 as the teacher and ResNet8$\times$4 as the student model. `N/A' denotes the baseline, which is the KD with prediction augmentation (PA). $gt^{\prime}$ is the new confidence value on the ground-truth index. $max^{\prime}$ is the new maximum confidence.}
\label{table_comparison_logit}
\centering
\begin{tabular}{cccc}
\hline
Method & Implementation & Top-1 (\%) & $\Delta$
\\
\hline
N/A &  $gt^{\prime} = gt$ & 76.91 & \color{gray}(+0.00)
\\
LSR & $gt^{\prime} = (1-\epsilon), p_{\symbol{92} t}=\epsilon/C$ & 77.32 & (+0.41)
\\
EGA & $gt^{\prime} = n \times gt$ & 74.41 & \color{red}(-2.50) 
\\
EGR & $gt^{\prime} = gt / n$ & 74.32 & \color{red}(-2.59)
\\
GA & $gt^{\prime} = gt + (gt \times w)$ & 77.17 & (+0.26)
\\
MA & $gt^{\prime} = max + (max \times w)$ & 77.34 & (+0.43)
\\
SLD & $gt^{\prime} = max, max^{\prime} = gt$ & \textbf{77.69} & (+0.78)
\\
\hline
\end{tabular}
\end{table*}

\subsection{Discussion}
It has been commonly acknowledged that the prediction of neural networks with higher confidence is more likely to be correct. However, the highest value must be on the ground-truth index to improve the model's prediction. Despite the highest value, how natural the confidence remains an important question. In this subsection, we demonstrate empirical comparisons for analyzing the effects of logit processing.  Based on these analyses, we propose a simple and effective non-parametric swap processing scheme on logit to correct the model's prediction without changing the non-class distribution. In Table \ref{table_comparison_logit}, we present the comparisons.

\noindent\textbf{Extreme Values are Unnatural.}
First, we conduct an experiment with a straightforward logit processing called extreme ground-truth addition (EGA). It enormously adds the $gt^{\prime}$ by double ($n = 2$). Even though the model's prediction is correct, the performance result drops significantly. This indicates that every correct prediction is not necessarily natural. As an outcome, the distribution context of non-class indices significantly changed. Afterward, extreme ground-truth reduction (EGR) is exploited by reducing the $gt^{\prime}$ by half. However, the results obtained are nearly identical to those obtained by EGA and even more inferior. This shows that the ground-truth adjusting with extreme values may negatively impact the performance.

\noindent\textbf{Soft Addition for Prediction Correction.}
After we demonstrate that extreme values are unnatural, we introduce a soft addition for prediction correction. Ground-truth addition (GA) adds the $gt^{\prime}$ with a small value $w = 0.1$ of $gt$ to maintain the value of confidence in the natural scope. As a result, GA achieves better performance than the baseline. However, by some chance, GA possibly predicts wrong when the $gt^{\prime}$ is not the maximum. To address this problem, we propose maximum addition (MA) to guarantee that $gt^{\prime}$ is the maximum in the distribution and reduce a significant change to the rest. In this experiment, the soft addition improves the performance but changes the non-class distribution because of temperatures and $\texttt{softmax}$ function effects.

\noindent\textbf{Swapping Logits.}
Instead of exhaustively searching for the optimal $w$ value, we provide a new simple non-parametric solution. We propose SLD that swaps the $gt$ and $max$. We swap them because they tend to have high values and share similarities in the input data. Suppose we have a “cat” as a label and a “dog” as another index with maximum confidence; they have more similarities than a “cat” and a “truck”. By swapping them, the non-class distribution is not changed, and the correct prediction can be obtained at the same time. When the $gt$ is obtained on training with cross-entropy loss, it will be higher than the other classes. With this sense, the $max^{\prime}$ will not significantly reduce, causing it to contain correlated context. The results show that SLD leads to improvements compared to other strategies.

\noindent\textbf{Comparison with Label Smoothing.} 
LSR \cite{lsr} regularizes model training by leveraging the one-hot target to a fixed smooth distribution including $gt^{\prime}$ and non-target $p_{\symbol{92} t}$ with a weight $\epsilon$ and number of classes $C$. To some extent, SLD is similar to LSR in a way that it involves a virtual teacher via a logit processing scheme to align with ground truth when discrepancies occur. In the experiment, we replace our swap method for the pseudo-teacher with LSR with $\epsilon=0.1$. We observe that our method has better advantages since the teacher's non-target distribution still keeps meaningful context for the distillation. On the other hand, LSR excludes them with a uniform distribution.

\section{Experiments}
\subsection{Experimental Setup}

\noindent\textbf{Datasets.} We conduct experiments on CIFAR-100 \cite{cifar} and ImageNet \cite{imagenet} datasets. 1) CIFAR-100 is a well-known image classification dataset with a resolution of 32x32 pixels and 100 categories. It consists of 50,000 training and 10,000 validation images. 2) ImageNet is a challenging image classification dataset of 1000 categories. It contains around 1.3 million training and 50,000 validation images.

\noindent\textbf{Settings.} We experiment with various neural network architectures, including VGGs \cite{vgg}, ResNets \cite{resnet}, WideResNets (WRN) \cite{wrn}, MobileNets \cite{mobilenet1,mobilenet2}, and ShuffleNets \cite{shufflenet1, shufflenet2}. All extensive experiments are performed in two settings, \textit{i.e.}, homogeneous architectures, where models are in the same architecture type, and heterogeneous architectures, where models are in different architecture types.

\noindent\textbf{Baselines.} We compare SLD with other knowledge distillation methods. For feature-based distillation, we compare with FitNet \cite{fitnets}, AT \cite{at}, RKD \cite{rkd}, CRD \cite{crd}, OFD \cite{ofd}, ReviewKD \cite{reviewkd}, and CAT-KD \cite{catkd}. For logit-based distillation, we compare with KD \cite{kd}, DML \cite{dml}, TAKD \cite{takd}, DKD \cite{dkd}, MLKD \cite{mlkd}, WTTM \cite{wttm}, and LS-MLKD \cite{normkd}.

\noindent\textbf{Implementation Details.} We implement all standard networks and training procedures in PyTorch. The models are trained following MLKD's \cite{mlkd} training configurations, except for 480 training epochs on CIFAR-100. For the CIFAR-100, we train the model on 1 GPU for 240 epochs with a batch size of 64. The initial learning rate of 0.01 is used for MobileNets and ShuffleNets, and 0.05 for other series (\textit{e.g.}, VGGs, ResNets, WRNs). The weight decay and the momentum are set to 5e-4 and 0.9. For ImageNet, we train the model on 4 GPUs with a batch size of 128 and an initial learning rate of 0.2 with division by 10 for every 30 epochs. All experiment results are averaged over four runs.

\begin{table*}[ht]
\renewcommand{\arraystretch}{1.1}
\caption{CIFAR-100 results, homogeneous architecture. Top-1 accuracy (\%) is adopted as the evaluation metric. \textbf{bold} and \underline{underline} represent the best and the second best performance.}
\label{table_homogeneous}
\centering
\resizebox{\columnwidth}{!}{\begin{tabular}{cc|ccccccc}
\multirow{4}{*}{Method}
& \multirow{2}{*}{Teacher}
& ResNet56
& ResNet110
& ResNet32$\times$4
& WRN-40-2
& WRN-40-2
& VGG13
& \multirow{4}{*}{Avg}
\\
  & & 72.34 & 74.31 & 79.42 & 75.61 & 75.61 & 74.64 &
\\
  & \multirow{2}{*}{Student} & ResNet20 & ResNet32 & ResNet8$\times$4 &  WRN-16-2 & WRN-40-1 & VGG8 &
\\
  & & 69.06 & 71.14 & 72.50 & 73.26 & 71.98 & 70.36 &
\\
\hline
\multirow{5}{*}{Feature} & FitNet & 69.21 & 71.06 & 73.50 & 73.58 & 72.24 & 71.02 & 71.77
\\
& RKD & 69.61 & 71.82 & 71.90 & 73.35 & 72.22 & 71.48 & 71.73
\\
& CRD & 71.16 & 73.48 & 75.51 & 75.48 & 74.14 & 73.94 & 73.95
\\
& OFD & 70.98 & 73.23 & 74.95 & 75.24 & 74.33 & 73.95 & 73.78
\\
& ReviewKD & 71.89 & 73.89 & 75.63 & 76.12 & 75.09 & 74.84 & 74.58
\\
& CAT-KD & 71.62 & 73.62 & 76.91 & 75.60 & 74.82 & 74.65 & 74.54
\\
\hline
\multirow{6}{*}{Logit} & KD & 70.66 & 73.08 & 73.33 & 74.92 & 73.54 & 72.98 & 73.09
\\
& DML & 69.52 & 72.03 & 72.12 & 73.58 & 72.68 & 71.79 & 71.95
\\
& TAKD & 70.83 & 73.37 & 73.81 & 75.12 & 73.78 & 73.23 & 73.36
\\
& DKD & 71.97 & 74.11 & 76.32 & 76.24 & 74.81 & 74.68 & 74.69
\\
& MLKD & 72.19 & 74.11 & 77.08 & 76.63 & 75.35 & 75.18 & 75.09
\\
& WTTM & 71.92 & 74.13 & 76.06 & 76.37 & 74.58 & 74.44 & 74.58
\\
& LS-MLKD & \underline{72.33} & \underline{74.32} & \textbf{78.28} & \underline{76.95} & \underline{75.56} & \underline{75.22} & \underline{75.44}
\\
 & SLD & \textbf{72.67} & \textbf{74.57} & \underline{77.69} & \textbf{77.19} & \textbf{76.36} & \textbf{75.33} & \textbf{75.64}
 \\
 \hline
\end{tabular}
}
\end{table*}

\begin{table*}[h]
\renewcommand{\arraystretch}{1.1}
\caption{CIFAR-100 results, heterogeneous architecture. Top-1 accuracy (\%) is adopted as the evaluation metric. $^\dagger$ denotes reproduced by our implementation. \textbf{bold} and \underline{underline} represent the best and the second best performance.}
\label{table_heterogeneous}
\centering
\resizebox{\columnwidth}{!}{
\begin{tabular}{cc|cccccc}
\multirow{4}{*}{Method}
& \multirow{2}{*}{Teacher}
& ResNet32$\times$4
& WRN-40-2
& VGG13
& ResNet50
& ResNet32$\times$4
& \multirow{4}{*}{Avg}
\\
  & & 79.42 & 75.61 & 74.64 & 79.34 & 79.42 & 
\\
  & \multirow{2}{*}{Student} & ShuffleNet-V1 & ShuffleNet-V1 & MobileNet-V2 & MobileNet-V2 & ShuffleNet-V2 &
\\
  & & 70.50 & 70.50 & 64.60 & 64.60 & 71.82 & 
\\
\hline
\multirow{5}{*}{Feature} & FitNet & 73.59 & 73.73 & 64.14 & 63.16 & 73.54 & 69.63
\\
& RKD & 72.28 & 72.21 & 64.52 & 64.43 & 73.21 & 69.33
\\
& CRD & 75.11 & 76.05 & 69.73 & 69.11 & 75.65 & 73.13
\\
& OFD & 75.98 & 75.85 & 69.48 & 69.04 & 76.82 & 73.43
\\
& ReviewKD & 77.45 & 77.14 & 70.37 & 69.89 & 77.78 & 74.53
\\
& CAT-KD & \textbf{78.26} & 77.35 & 69.13 & \underline{71.36} & 78.41 & 74.90
\\
\hline
\multirow{6}{*}{Logit} & KD & 74.07 & 74.83 & 67.37 & 67.35 & 74.45 & 71.60
\\
& DML & 72.89 & 72.76 & 65.63 & 65.71 & 73.45 & 70.09
\\
& TAKD & 74.53 & 75.34 & 67.91 & 68.02 & 74.82 & 72.12
\\
& DKD & 76.45 & 76.70 & 69.71 & 70.35 & 77.07 & 74.69
\\
& MLKD & 77.18 & \underline{77.44} & 70.57 & 71.04 & 78.44 & 74.93
\\
& WTTM & 74.37 & 75.42 & 69.59 & 69.16 & 76.55 & 73.02
\\
& LS-MLKD & \text{ }77.29$^\dagger$ & \text{ }77.57$^\dagger$ & \textbf{70.94} & 71.19 & \underline{78.76} & \underline{75.15}
\\
 & SLD & \underline{77.48} & \textbf{77.69} & \underline{70.76} & \textbf{71.48} & \textbf{78.82} & \textbf{75.24} 
 \\
 \hline
\end{tabular}
}
\end{table*}

\begin{table*}[h]
\renewcommand{\arraystretch}{1.1}
\caption{ImageNet results. \textbf{bold} and \underline{underline} represent the best and the second best performance. Top-1 and top-5 accuracies are reported.}
\label{table_imagenet}
\centering
\begin{tabular}{cc|cc|cc}
& & Top-1 & Top-5 & Top-1 & Top-5
\\
\hline
\multirow{4}{*}{Method} & \multirow{2}{*}{Teacher} & \multicolumn{2}{c|}{ResNet34} & \multicolumn{2}{c}{ResNet50}
\\
& & 73.31 & 91.42 & 76.16 & 92.86
\\
& \multirow{2}{*}{Student} & \multicolumn{2}{c|}{ResNet18} & \multicolumn{2}{c}{MobileNetV1}
\\
& & 69.75 & 89.07 & 68.87 & 88.76
\\
\hline
\multirow{4}{*}{Feature} & AT & 70.69 & 90.01 & 69.56 & 89.33
\\
& OFD & 70.81 & 89.98 & 71.25 & 90.34
\\
& CRD & 71.17 & 90.13 & 71.37 & 90.41
\\
& ReviewKD & 71.61 & 90.51 & 72.56 & 91.00
\\
& CAT-KD & 71.26 & 90.45 & 72.24 & 91.13
\\
\hline
\multirow{4}{*}{Logit} & KD & 70.66 & 89.88 & 68.58 & 88.98
\\
& TAKD & 70.78 & 90.16 & 70.82 & 90.01
\\
& DKD & 71.70 & 90.41 & 72.05 & 91.05
\\
& MLKD & 71.90 & 90.55 & 73.01 & 91.42
\\
& LS-MLKD & \underline{72.08} & \underline{90.74} & \underline{73.22} & \underline{91.59}
\\
& SLD & \textbf{72.15} & \textbf{90.90} & \textbf{73.27} & \textbf{91.65} 
\\
\hline
\end{tabular}
\end{table*}

\subsection{Experimental Results}

\noindent\textbf{Results on CIFAR-100.}
The experiment results on CIFAR-100 are shown in Table \ref{table_homogeneous} and \ref{table_heterogeneous}. SLD consistently outperforms the previous works, including the feature-based and logit-based knowledge distillation approaches. Compared with MLKD as our baseline, SLD achieves better performance on average. These results show the effectiveness of the proposed SLD in dealing with teacher and student models with homogeneous and heterogeneous architectures. Meanwhile, most logit-based distillation methods remain a trade-off with the feature-based distillation method, specifically with heterogeneous architectures.

\noindent\textbf{Results on ImageNet.} 
SLD is compared with the previous methods in two settings, homogeneous and heterogeneous architectures on ImageNet, as shown in Table \ref{table_imagenet}. It is worth mentioning that SLD performs better than the other distillation methods in Top-1 and Top-5 accuracy.

\begin{table*}[h]
\renewcommand{\arraystretch}{1.1}
\caption{Ablation study. The experiments are implemented on CIFAR-100.}
\label{table_ablation_study}
\centering
\resizebox{\columnwidth}{!}{\begin{tabular}{ccc|cccccc}
\multicolumn{3}{c|}{Teacher} & Res56 & Res110 & Res32$\times$4 & WRN-40-2 & WRN-40-2 & VGG13
\\
 & & & 72.34 & 74.31 & 79.42 & 75.61 & 75.61 & 74.64
\\
\multicolumn{3}{c|}{Student} & Res20 & Res32 & Res8$\times$4 & WRN-16-2 & WRN-40-1 & VGG8
\\
 & & & 69.06 & 71.14 & 72.50 & 73.26 & 71.98 & 70.36
\\
\hline
 $\mathcal{L}_{TS}$ & $\mathcal{L}_{SS}$ & PA & \textcolor{gray}{70.66} & \textcolor{gray}{73.08} & \textcolor{gray}{73.33} & \textcolor{gray}{74.92} & \textcolor{gray}{73.54} & \textcolor{gray}{72.98}
\\
\cmark & & & 71.03 & 73.35 & 74.84 & 75.52 & 74.07 & 73.46
\\
\cmark & \cmark & & 71.55 & 73.56 & 75.15 & 75.87 & 74.15 & 73.70
\\
\cmark & \cmark & \cmark & \textbf{72.67} & \textbf{74.57} & \textbf{77.69} & \textbf{77.19} & \textbf{76.36} & \textbf{75.33}
\\
\hline

\multicolumn{3}{c|}{Teacher} & Res32$\times$4 & WRN-40-2 & VGG13 & Res50 & Res32$\times$4 & WRN-40-2 
\\
 & & & 79.42 & 75.61 & 74.64 & 79.34 & 79.42 & 75.61
\\
\multicolumn{3}{c|}{Student} & ShuV1 & ShuV1 & MV2 & MV2 & ShuV2 & VGG8 
\\
 & & & 70.50 & 70.50 & 64.60 & 64.60 & 71.82 & 70.36 
\\
\hline
 $\mathcal{L}_{TS}$ & $\mathcal{L}_{SS}$ & PA & \textcolor{gray}{74.07} & \textcolor{gray}{74.83} & \textcolor{gray}{67.37} & \textcolor{gray}{67.35} & \textcolor{gray}{74.45} & \textcolor{gray}{73.55}
\\
\cmark & & & 74.21 & 75.28 & 67.85 & 68.37 & 74.93 & 73.68 
\\
\cmark & \cmark & & 74.38 & 75.61 & 67.90 & 68.61 & 75.12 & 73.91
\\
\cmark & \cmark & \cmark & \textbf{77.48} & \textbf{77.69} & \textbf{70.76} & \textbf{71.48} & \textbf{78.82} & \textbf{75.40}
\\
\hline

\end{tabular}}
\end{table*}

\subsection{Analysis}

\noindent\textbf{Ablation Study.}
To dissect the effect of each component, we conduct an ablation study, removing individual elements from SLD to conventional KD. As shown in Table \ref{table_ablation_study}, $\mathcal{L}_{TS}$ and $\mathcal{L}_{SS}$ consistently lead to a non-trivial improvement in the performance of conventional KD. When the $\mathcal{L}_{TS}$ and $\mathcal{L}_{SS}$ meet the PA, the best performance in Top-1 accuracy is obtained. This proves that combining all elements is imperative, and swapping strategy effectively improves performance.

\begin{table*}[h]
\renewcommand{\arraystretch}{1.2}
\caption{Performance gap between teacher and student model. CIFAR-100 is used for experiments with homogeneous architectures. Top-1 accuracy as the evaluation metric. Negative values indicate when the student model outperforms the teacher model.}
\label{table_teachergap}
\centering
\begin{tabular}{c|cccccc|c}
Teacher & 72.34 & 74.31 & 79.42 & 75.61 & 75.61 & 74.64 & 75.32 (Avg)
\\
\hline
Student & 69.06 & 71.14 & 72.50 & 73.26 & 71.98 & 70.36 & 71.38 (Avg)
\\
Gap & 3.28 & 3.17 & 6.92 & 2.35 & 3.63 & 4.28 & 3.94 (Avg)
\\
\hline
SLD (w/o $\mathcal{L}_{SS}$) & 71.92 & 73.98 & 77.37 & 76.50 & 75.21 & 75.15 & 75.02 (Avg)
\\
Gap & 0.42 & 0.33 & 2.05 & -0.89 & 0.40 & -0.51 & 0.30 (Avg)
\\
\hline
SLD & 72.67 & 74.57 & 77.69 & 77.19 & 76.36 & 75.33 & 75.64 (Avg)
\\
Gap & \textbf{-0.33} & \textbf{-0.26} & 1.73 & \textbf{-1.58} & \textbf{-0.75} & \textbf{-0.69} & \textbf{-0.32} (Avg)
\\
\hline
\end{tabular}
\end{table*}

\noindent\textbf{Comparison with Teacher Model.}
To evaluate the quality of knowledge distillation techniques, we compare the performance gap between the teacher and student model shown in Table \ref{table_teachergap}. It is observed that the student model's performance with SLD can achieve better performance than the teacher model's, which is marked by a gap with a negative value on average. As discussed before, we hypothesized that the $\mathcal{L}_{SS}$ plays an important role as a pseudo-teacher model to retrieve the information independently outside of the base teacher model, causing an improvement over the teacher's accuracy. Therefore, we demonstrate the average performance of the student model with SLD (w/o $\mathcal{L}_{SS}$) and find that $\mathcal{L}_{SS}$ is an effective component to acquire the external information.

\begin{table*}[h]
\renewcommand{\arraystretch}{1.1}
\caption{Loss scheduling for $\mathcal{L}_{SS}$. The experiments are implemented on CIFAR-100.}
\label{table_schedule}
\centering
\resizebox{\columnwidth}{!}{
\begin{tabular}{c|cccccc}
Teacher & Res56 & Res110 & Res32$\times$4 & WRN-40-2 & WRN-40-2 & VGG13
\\
& 72.34 & 74.31 & 79.42 & 75.61 & 75.61 & 74.64
\\
Student & Res20 & Res32 & Res8$\times$4 & WRN-16-2 & WRN-40-1 & VGG8
\\
& 69.06 & 71.14 & 72.50 & 73.26 & 71.98 & 70.36
\\
\hline
\textcolor{gray}{w/o $\mathcal{L}_{SS}$} & \textcolor{gray}{71.92} & \textcolor{gray}{73.98} & \textcolor{gray}{77.37} & \textcolor{gray}{76.50} & \textcolor{gray}{75.21} & \textcolor{gray}{75.15}
\\
\xmark & 72.55 & 74.22 & 77.40 & 76.41 & 75.72 & 75.29
\\
\cmark & \textbf{72.67} & \textbf{74.57} & \textbf{77.69} & \textbf{77.19} & \textbf{76.36} & \textbf{75.33} 
\\
$\Delta$ & (+0.12) & (+0.35) & (+0.29) & (+0.78) & (+0.64) & (+0.04)
\\
\\[-2.5 ex] \hline \\[-2.5 ex]
Teacher & Res32$\times$4 & WRN-40-2 & VGG13 & Res50 & Res32$\times$4 & WRN-40-2
\\
& 79.42 & 75.61 & 74.64 & 79.34 & 79.42 & 75.61
\\
Student & ShuV1 & ShuV1 & MV2 & MV2 & ShuV2 & VGG8 
\\
& 70.50 & 70.50 & 64.60 & 64.60 & 71.82 & 70.36
\\
\hline
\textcolor{gray}{w/o $\mathcal{L}_{SS}$} & \textcolor{gray}{76.96} & \textcolor{gray}{77.32} & \textcolor{gray}{70.39} & \textcolor{gray}{70.75} & \textcolor{gray}{78.34} & \textcolor{gray}{75.15}
\\
\xmark & 74.07 & 76.45 & 70.61 & 65.98 & 78.10 & 75.12
\\
\cmark & \textbf{77.48} & \textbf{77.69} & \textbf{70.76} & \textbf{71.48} & \textbf{78.82} & \textbf{75.40}
\\
$\Delta$ & (+3.41) & (+1.24) & (+0.15) & (+5.50) & (+0.72) & (+0.28)
\\
\hline
\end{tabular} }
\end{table*}

\noindent\textbf{How to Integrate Teachers.}
In Table \ref{table_schedule}, we observe that by directly integrating $\mathcal{L}_{SS}$ with $\mathcal{L}_{TS}$, the performance drops up to 5.50$\%$, particularly in heterogeneous architecture. Intuitively, this is because the student distribution conflicts with the teacher's from the beginning of the training. For example, the VGG13-MV2 is less sensitive than the Res50-MV2 in performance results produced by different teachers' models. Surprisingly, the integration can be simply solved by scheduling the $\mathcal{L}_{SS}$ to boost the performance results of all models. This approach mimics human learning by starting with simple tasks given by the teacher and gradually learning harder ones by self-study once the developed student has absorbed the knowledge from the teacher.

\begin{table*}
\renewcommand{\arraystretch}{1.1}
\caption{Combination with other distillation methods. CIFAR-100 is used for experiments. Top-1 accuracy as the evaluation metric.}
\label{table_combination}
\centering
\resizebox{\columnwidth}{!}{
\begin{tabular}{c|cc|cc|cc|cc|cc}
Model & RKD & \textbf{+SLD} & ReviewKD & \textbf{+SLD} & DKD & \textbf{+SLD} & MLKD & \textbf{+SLD} & LS-MLKD & 
 \textbf{+SLD}
\\
\hline
WRN-40-2, VGG8 & 71.22 & \textbf{75.82} & 74.91 & \textbf{75.89} & 74.20 & \textbf{75.71} & 74.88 & \textbf{75.70} & 75.42 & \textbf{76.43}
\\
Res32$\times$4, Shu-V1 & 72.28 & \textbf{77.50} & 77.45 & \textbf{77.90} & 76.45 & \textbf{78.67} & 77.18 & \textbf{77.98} & 77.29 & \textbf{78.73}
\\
Res32$\times$4, Res8$\times$4 & 71.90 & \textbf{77.86} & 75.63 & \textbf{77.93} & 76.32 & \textbf{77.74} & 77.08 & \textbf{77.82} & 78.28 & \textbf{78.66}
\\

\end{tabular}
}
\end{table*}

\noindent\textbf{Combination with Other Distillation Methods.}
Existing work \cite{mlkd} observes the combination only with feature-based distillation. As shown in Table \ref{table_combination}, we integrate the proposed method with both logit and feature-based distillation to validate the generalizability. We demonstrate with three settings, \textit{i.e.}, heterogeneous with different layers and structures (column 1), heterogeneous with different structures (column 2), and homogeneous (column 3). The experimental results show that combining other distillation methods with ours further improves the baselines' performance by a large margin. This shows that our approach is easy to blend with other distillation methods since our approach is based on instance-level relations.

\begin{table*}[h]
\renewcommand{\arraystretch}{1.1}
\caption{Performance of SLD with different temperature scaling of prediction augmentation.}
\label{table_pa}
\centering
\begin{tabular}{c|cccccc}
$T$ & 2-5 & 1-5 & 2-6 & 1-6 & 2-7 & 1-7
\\
\hline
median & 3.5 & 3 & 4 & 3.5 & 4.5 & 4
\\
$k$ & 4 & 5 & 5 & 6 & 6 & 7
\\
\hline
Top-1 (\%) & 77.16 & 77.44 & 77.52 & \textbf{77.69} & 77.34 & 77.58
\end{tabular}
\end{table*}

\noindent\textbf{Temperature Scaling.}
To verify the performance consistency, we report the student accuracy (\%) with different $T$ on CIFAR-100 in Table \ref{table_pa}. ResNet32$\times$4 and ResNet8$\times$4 are set as the teacher and the student, respectively. In the \cite{mlkd}, the ideal $T$ is from 2 to 6 with a median of 4. We observe that the swap method is more beneficial with the prediction augmentation than MLKD (77.08\% v.s. 77.52\%). In these experiments, the model performs best with $T$ = [1.0, 2.0, 3.0, 4.0, 5.0, 6.0]. This is because the low temperature of 1.0 is vital for prediction augmentation, enabling more decisive learning for the swap mechanism. Henceforth, we set $T$ with a median of 3.5 and a temperature length $k$ of 6 in the experiments.

\begin{figure}[h]
\includegraphics[scale=0.6]{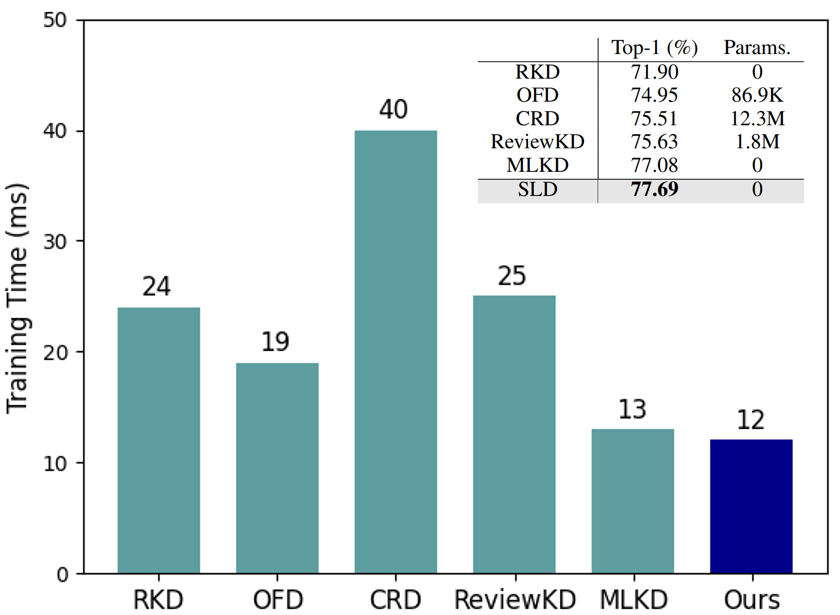}
\centering
\caption{Training time (per batch). We set ResNet32$\times$4 as the teacher model and ResNet8$\times$4 as the student model on CIFAR-100. The table shows the accuracy and the number of additional parameters.}
\label{fig:training_time}
\end{figure}

\noindent\textbf{Training Efficiency.} We assess the training costs (\textit{e.g.}, time and extra parameters) of SLD to measure its efficiency shown in Fig. \ref{fig:training_time}. SLD achieves the best performance with less training time and no extra parameters. This suggests that the training process is efficient with merely logit information. Compared with MLKD, SLD is simpler and more effective since the computation comes from instance-level relations similar to the original KD \cite{kd}. Besides, feature-based distillations compute more time in distilling features from the intermediate layers (additional operations and modules).

\begin{figure}[h]
\includegraphics[scale=0.45]{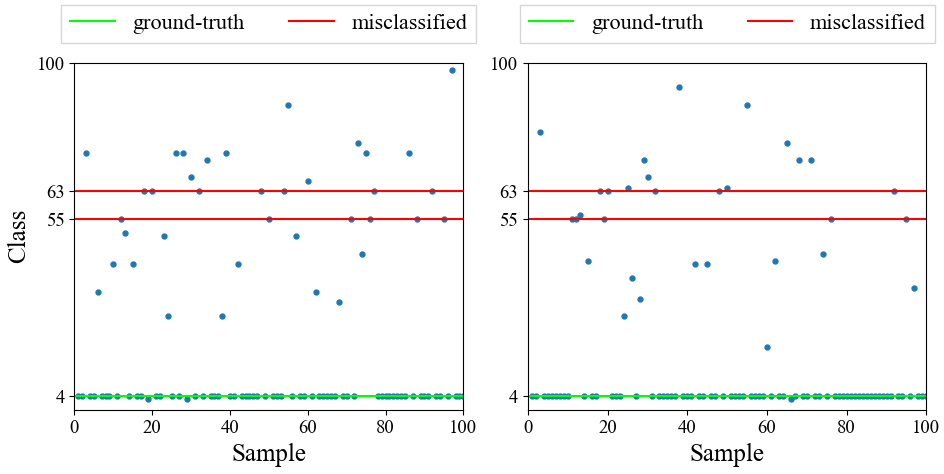}
\centering
\caption{Prediction distribution of KD (left) and KD with swap method (right). Class 4 (beaver) is the ground-truth, class 55 (otter), and class 63 (porcupine).}
\label{fig:pred_distribution}
\end{figure}

\noindent\textbf{Prediction Distribution.}
To verify the effects of the swap mechanism against the test sample, we show the prediction distribution in Fig. \ref{fig:pred_distribution}. Specifically, we compare the KD and KD+$\mathcal{L}_{TS}$+$\mathcal{L}_{SS}$ results with ResNet32$\times$4 - ResNet8$\times$4 models on CIFAR100. Our method predicts 68, 5, and 5 for `beaver' as a target, `otter', and `porcupine', respectively. In contrast, KD predicts 61, 6, and 7, respectively. This demonstrates that the swap methods enable the student model to predict more accurately than the KD while the misclassification of similar semantic samples can be reduced.

\begin{table*}[h]
\renewcommand{\arraystretch}{1.1}
\caption{Differences of student and teacher logits' correlation matrices.}
\label{table_correlation}
\centering
\begin{tabular}{c|cc}
Method & Max $\downarrow$ & Mean $\downarrow$
\\
\hline
KD & 1.078 & 0.091
\\
MLKD & 0.858 & 0.080
\\
SLD & \textbf{0.837} & \textbf{0.078}
\\
\end{tabular}
\end{table*}

\noindent\textbf{Correlation Matrices of Student and Teacher.}
In Table \ref{table_correlation}, we present differences in correlation matrices of the student and teacher logits. We take ResNet32$\times$4 as the teacher and ResNet8$\times$4 as the student on CIFAR-100. Compared with KD and MLKD, the student model learns from the teacher model better, as indicated by the average distance between the student and teacher model.

\begin{table*}[h]
\renewcommand{\arraystretch}{1.1}
\caption{Performance of various swap scenarios conducted on CIFAR-100 with ResNet32$\times$4 as the teacher and ResNet8$\times$4 as the student model.}
\label{table_multiswap}
\centering
\begin{tabular}{c|ccc}
Method & Swap Process & Action & Acc (\%)
\\
\hline
w/o $\mathcal{L}_{SS}$ & top-1 & Single & 77.37
\\
w/o $\mathcal{L}_{SS}$ & top-2 $\rightarrow$ top-1 & Multiple & 77.41
\\
w/o $\mathcal{L}_{SS}$ & top-3 $\rightarrow$ top-2 $\rightarrow$ top-1 & Multiple & 77.59
\\
SLD & top-1 & Single & \textbf{77.69}
\\
SLD & top-2 $\rightarrow$ top-1 & Multiple & 77.31
\\
SLD & top-3 $\rightarrow$ top-2 $\rightarrow$ top-1 & Multiple & 77.53
\end{tabular}
\end{table*}

\noindent\textbf{Does Multiple Swap Work?}
In Table \ref{table_multiswap}, we conduct an experiment to explore whether the swap mechanism could be extended to multiple processes rather than a single one. For example, we swap the target value with the top-2 prediction and continuously swap the target value with the top-1 prediction. This multiple swapping also applies to top-3 predictions. From the experiments, we observe that the multiple swap operation works well on the swapped teacher. Since the single swap of SLD still performs the best and yet it is simple, in this paper, we exploit the use of a single swap on bi-level teachers.

\begin{table*}[h]
\renewcommand{\arraystretch}{1.1}
\caption{Experimental results of conditional swap of teachers in Top-1 (\%) on CIFAR-100. $\alpha$ is a swapping threshold where $\alpha = abs(\max(softmax(p_k)) - softmax(p_k)[t])$.}
\label{table_conditionswap}
\centering
\begin{tabular}{c|c|ccccc}
Model & $\alpha$ & 0 & 0.25 & 0.50 & 0.75 & 1
\\
\hline
\multirow{2}{*}{w/o $\mathcal{L}_{SS}$} & Less than $\alpha$ & - & 77.41 & \textbf{77.68} & 77.48 & 77.37
\\
 & More than $\alpha$ & 77.37 & 77.29 & 77.13 & 77.17 & -
\\
\hline
\multirow{2}{*}{SLD} & Less than $\alpha$ & - & 77.59 & 77.53 & \textbf{77.97} & 77.69
\\
 & More than $\alpha$ & 77.69 & 77.19 & 77.08 & 77.27 & -
\\
\end{tabular}
\end{table*}

\noindent\textbf{Effects of Conditional Swap.}
In the previous section, the SLD method improves prediction by leveraging the assumption that the non-target class with the highest confidence shares a high similarity with the target class (\textit{e.g.,} “cat” and “dog”). What is the impact of swapping if the prediction is semantically different from the target class (\textit{e.g.,} “cat” and “truck”)? To verify this, we first obtain $\alpha$, which is a threshold computing the highest and the target confidence of the softmax of the prediction separately, and then find the absolute difference between the two results. Specifically, we only swap the logits based on the $\alpha$ condition. The higher the $\alpha$, the greater the divergence between the prediction confidence and the target confidence. In Table \ref{table_conditionswap}, we find that if the discrepancy between the prediction confidence and the target confidence is small, it leads to a better model performance.

\begin{table*}[h]
\renewcommand{\arraystretch}{1.1}
\caption{Experimental results demonstrating the impact of the pseudo-teacher on the student model’s performance on CIFAR-100. The PA includes an unswapped teacher.}
\label{table_studentswap_error}
\centering

\begin{tabular}{l|ccc}
Model & PA & PA + $\mathcal{L}_{SS}$ & SLD
\\
\hline
T: WRN-40-2, S: VGG8 & 75.11 & 75.25 (+0.14) & \textbf{75.40}
\\
T: ResNet32$\times$4, S: Shu-V1 & 76.85 & 77.31 (+0.46) & \textbf{77.48}
\\
T: ResNet32$\times$4, S: ResNet8$\times$4 & 76.91 & 77.24 (+0.33) & \textbf{77.69}
\\
\end{tabular}

\end{table*}

\noindent\textbf{Impact of Pseudo-Teacher with Vanilla Teacher.}
If the teacher model potentially generates inaccurate outputs for specific classes, relying on a pseudo-teacher may reinforce the prediction errors made by the student. To validate this, we conduct an experiment shown in Table \ref{table_studentswap_error}. We can see that even without swapping the teacher, the pseudo-teacher contributes to improving the performance of the student model. This is because the pseudo-teacher corrects the student's prediction regardless of the potential of the teacher's errors.

\noindent\textbf{Image Visualization.}
We present a visualization with the teacher model as ResNet32$\times$4 and the student model as ResNet8$\times$4 on CIFAR-100. We show the misclassified samples of original KD and MLKD that SLD correctly classifies in both ways. For example, the “hamster” is classified as the “mouse” and vice versa in the KD and MLKD cases. In contrast, SLD correctly predicts both samples. These verify that SLD enjoys the prediction improvement among samples with similar features.

\section{Conclusion}

To optimize the use of logit outputs, this paper reformulates the conventional KD to SLD, integrating the prediction augmentation and the novel logit swap mechanism to correct the prediction without changing the non-class distribution at several logit softness levels. As discussed before, we demonstrate the drawbacks when the logit processing is unnatural, resulting in sub-optimal performance. To overcome this limitation, SLD swaps the misclassified target with the non-target with the highest confidence of the logit distribution. Intuitively, the swap mechanism can guide the student to pay more attention to the target confidence instead of the misclassified non-target to achieve optimal performance. More importantly, the swap mechanism cooperates well with the teacher and student logits that provide independent information. Extensive experiments prove the effectiveness and improvement of our approach compared with the previous KD methods.

\backmatter

\bmhead{Acknowledgements}
This work is partially supported by the National Science and Technology Council (NSTC), Taiwan, under Grants: NSTC-112-2628-E-002-033-MY4, NSTC-113-2634-F-002-001-MBK, and NSTC-112-2221-E-A49-089-MY3 and was financially supported in part by the Center of Data Intelligence: Technologies, Applications, and Systems, National Taiwan University (Grants: 114L900901/114L900902/114L900903), from the Featured Areas Research Center Program within the framework of the Higher Education Sprout Project by the Ministry of Education, Taiwan.

\bibliography{sn-bibliography}

\end{document}